\documentclass[preprint,12pt,authoryear]{elsarticle}
\usepackage{natbib}
\usepackage{amsmath}
\usepackage{booktabs} 
\usepackage{caption} 
\usepackage{subcaption} 
\usepackage{graphicx}
\usepackage{pgfplots}
\usepackage[all]{nowidow}
\usepackage[utf8]{inputenc}
\usepackage{tikz}
\usetikzlibrary{er,positioning,bayesnet}
\usepackage{multicol}
\usepackage{booktabs}
\usepackage{algpseudocode,algorithm,algorithmicx}
\usepackage{hyperref}
\usepackage{tabularx}
\usepackage{rotating}
\usepackage{graphicx}
\usepackage{adjustbox}
\usepackage{graphicx}
\usepackage[inline]{enumitem} 
\usepackage{longtable} 
\usepackage{pdflscape} 
\usepackage{afterpage}
\usepackage{array}
\usepackage[T1]{fontenc}
\newcolumntype{P}[1]{>{\centering\arraybackslash}p{#1}}
\definecolor{blue}{HTML}{1F77B4}
\definecolor{orange}{HTML}{FF7F0E}
\definecolor{green}{HTML}{2CA02C}

\pgfplotsset{compat=1.14}

\journal{arXiv}

\begin{document}
\begin{frontmatter}

\title{The Application of Machine Learning Techniques for Predicting Results in Team Sport: A Review}

\author[1]{Rory Bunker\corref{mycorrespondingauthor}}
\address[1]{Nagoya Institute of Technology. Gokisocho, Showa Ward, Nagoya, Aichi, 466-8555, Japan}
\ead{rorybunker@gmail.com}
\cortext[mycorrespondingauthor]{Corresponding author}

\author[2]{Teo Susnjak}
\address[2]{Massey University. Massey University East Precinct, Dairy Flat Highway (SH17), 0632, New Zealand}
\ead{t.susnjak@massey.ac.nz}

\begin{abstract}
Over the past two decades, Machine Learning (ML) techniques have been increasingly utilized for the purpose of predicting outcomes in sport. In this paper, we provide a review of studies that have used ML for predicting results in team sport, covering studies from 1996 to 2019. We sought to answer five key research questions while extensively surveying papers in this field. This paper offers insights into which ML algorithms have tended to be used in this field, as well as those that are beginning to emerge with successful outcomes. Our research highlights defining characteristics of successful studies and identifies robust strategies for evaluating accuracy results in this application domain. Our study considers accuracies that have been achieved across different sports and explores the notion that outcomes of some team sports could be inherently more difficult to predict than others. Finally, our study uncovers common themes of future research directions across all surveyed papers, looking for gaps and opportunities, while proposing recommendations for future researchers in this domain.

\begin{keyword}
Machine Learning \sep Sport Result Prediction \sep Literature Review \sep Expert System \sep Team Sport


\end{keyword}
\end{abstract}

\end{frontmatter}

\section{Introduction}

Sport result prediction is an interesting and challenging problem due to the inherently unpredictable nature of sport, and the seemingly endless number of potential factors that can affect results. Indeed, it is this unpredictable nature that is one of the main reasons that people enjoy sport. Despite its difficulty, attempting to predict the outcome of sport is something that is of interest to many different stakeholders, and this interest has increased with sport data becoming increasingly available online, as well as with the advent of online sports betting. Sport experts or former players are often asked their predictions for forthcoming matches, which then appear online or in newspapers. Other stakeholders who are interested in predicting sport results include bookmakers and sports betting platforms, as well as gamblers who bet on match results, or for certain events occurring within a particular match. Note that the prediction of a specific event occurring within a match is a more general task, and may include events other than the final result, such as the occurrence of one particular event in a match, e.g., a goal being scored by a specific player in a match. The focus of this review is on the prediction of final results. Sport result prediction is also potentially useful to players, team management and performance analysts in identifying the most important factors that help to achieve winning outcomes, upon which appropriate tactics can be identified.

In the academic context, the problem of sport result prediction has been considered in the Statistics and Operations Research literature for some time, however the application of Machine Learning (ML) techniques to the problem is a more recent area of research. The first study in this domain was published in 1996, and the topic has attracted increased research interest since then. The scope of this review is on papers that have made use of at least one ML technique for the problem of sport result prediction. For the sake of brevity, we focus this review on prediction of team sports. The prediction of outcomes in individual or non-team sports has also been studied, e.g., in Horse Racing \citep{davoodi2010horse}, Swimming \citep{edelmann2002modeling}, Golf \citep{wiseman2016using}, Tennis \citep{somboonphokkaphan2009tennis} and Javelin \citep{maszczyk2014application}.
Fuzzy logic and rule-based methods, which are not the focus of this review, have also been applied to sport result prediction, e.g., by \cite{rotshtein2005football}, \cite{tsakonas2002soft} and \cite{min2008compound}. Deep Learning has shown potential very recently, with studies by \cite{chen2020using} and \cite{rudrapal2020deep}, but is not the focus of this review.

To our knowledge, there have been two prior review papers that considered ML for sport result prediction, by \cite{haghighat2013review} and, more recently, by \cite{keshtkar2019sports}. The present review is justified on the grounds that it draws out more critical insight obtained from previous studies as covered by Haghighat et al. \cite{haghighat2013review}, and it expands on the recent review by \cite{keshtkar2019sports} by providing a more comprehensive coverage of surveyed papers. 

\citet{schumaker2010sports} stated that Artificial Neural Networks (ANNs) are one of the most predominant ML techniques used in sports. Indeed, in the course of our review, we found that a number of studies, particularly early ones, made use of ANNs as their sole predictive model. This leads us to investigate whether the evidence in the literature suggested that ANNs actually perform better than other ML models in practice for problem of sport result prediction. One of the aims of this review is to consider this, as well as four additional key research questions, which are listed as follows:

\begin{enumerate}
\item Does the evidence from the literature suggest that ANNs produce better performance in terms of predictive accuracy than other ML models?
\item What are some of the defining characteristics that can be drawn from successful studies?
\item Are the results of some sports inherently more difficult to predict than in others?
\item What are the best ways to evaluate accuracy results in this application domain?
\item What are the common themes of future research directions that can be identified from surveyed papers?
\end{enumerate}

Our contribution lies in an extensive and up-to-date literature review which collates key studies carried out in the emerging field of predictive analytics in sports. We highlight which algorithms have been preferred by practitioners and have been at the forefront of this field. This review identifies what have been some of the key drivers of successful studies with respect to how machine learning was applied, and how those elements have contributed to high predictive accuracies. This review makes a meaningful contribution in identifying future research trends and opportunities in this field, while combining all the findings into a useful set of recommendations for other practitioners in this domain. 

The remainder of this paper is structured as follows. In section \ref{method} we outline our methodological approach and inclusion criteria for this study. Then, in section \ref{lit-review}, we review the literature in machine learning for sport result prediction and also present tabular summaries of studies by sport. We address the aforementioned key research questions, and in doing so, provide a discussion of these results, as well as potential limitations of our review, in section \ref{results-discussion}. We conclude in section \ref{limitations-conclusions}.

\section{Methodology}
\label{method}

\begin{figure}[htb]
 \centering\includegraphics[width=1\linewidth]{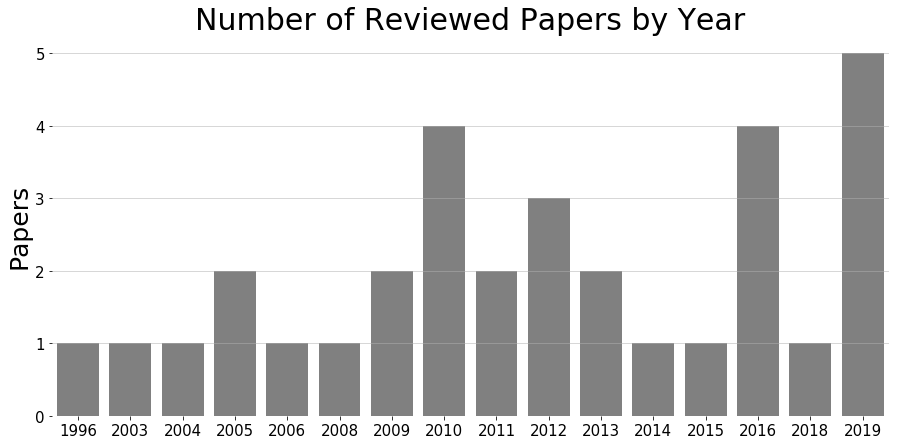}
 \caption{Number of papers by year considered in this review.}
 \label{papers_by_year}
 \end{figure}

\begin{figure}[htb]
 \centering\includegraphics[width=1\linewidth]{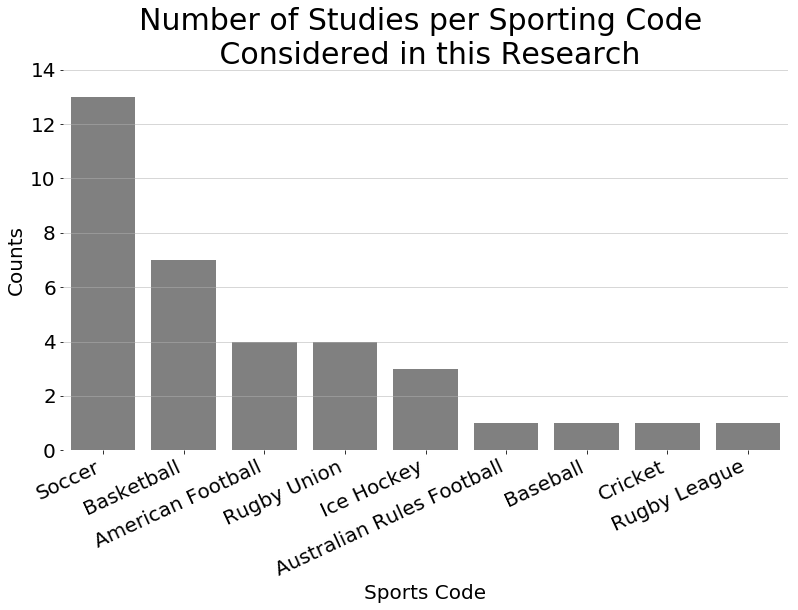}
 \caption{Number of sporting codes studied between 1996 - 2019 for the purposes of predictive modeling which were considered in this paper.}
 \label{sport_code_counts}
 \end{figure}

A total of 31 papers were considered in the present study. These were collected from the period of 1996 to 2019. The distribution of papers by year of publication can be seen in Figure \ref{papers_by_year}, showing a growing level of interest in this field from 2009 onward. These papers covered nine separate sports. Given that some papers studied multiple sports, a total of 35 sport studies were conducted. The specific sports and the distribution of the number of studies they received can be seen in Figure \ref{sport_code_counts}.

\paragraph{Inclusion Criteria:} Our review includes papers that incorporated at least one ML technique in their experimentation and therefore the review did not consider statistical, mathematical or operational research studies related to sport result prediction that do not use these techniques. Our focus is on the application of ML - fuzzy methods and Deep Learning methods are not the focus of this review. We also restricted our review to team sports. Papers generally needed to be published in peer-reviewed academic journals or conference proceedings in order to be included.
 
\begin{figure}[htb]
 \centering\includegraphics[width=1\linewidth]{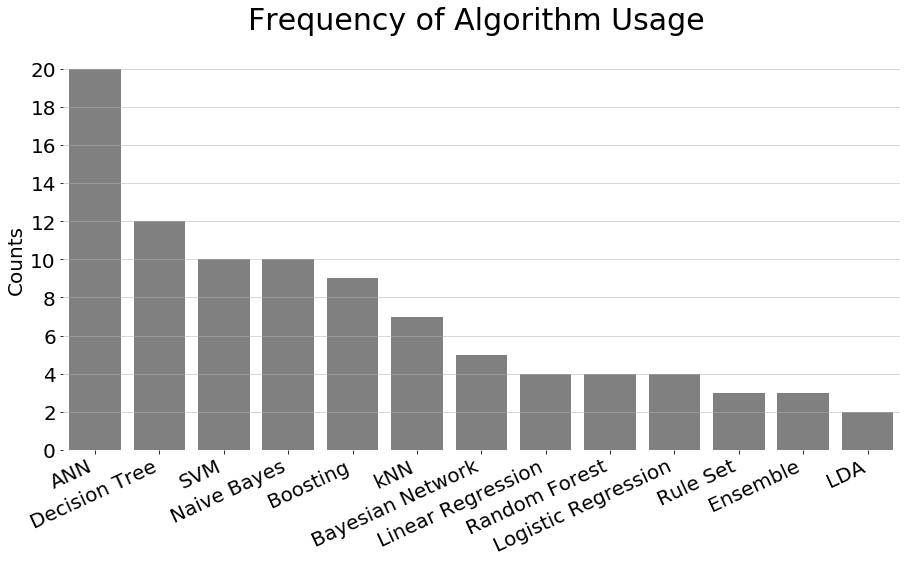}
 \caption{Histogram of usage patterns of various algorithms covered in this research. }
 \label{algorithm_usage}
 \end{figure}

\paragraph{Algorithm Grouping:} Machine Learning algorithms were normalized by grouping them into families of algorithms in order to identify trends in their usage patterns. All variants of ANN such as Back Propagation (BP), Feed-Forward, as well as Self Organising Maps and Long Short Term Memory ANNs were clustered under the same umbrella of algorithms. CART, C4.5 and WEKA's J48 algorithms were grouped under the Decision Tree family of methods. RIPPER, FURIA and ZeroR were merged into the Rule Sets category, while the Local Weighted Learning (LWL) algorithm was merged with k-Nearest-Neighbors (kNN). Additionally, AdaBoost, XGBoost, LogitBoost, RobustBoost, and RDN-Boost were clustered in the Boosting category, while Sequential Minimal Optimization (SMO) was combined with Support Vector Machines (SVMs). Methods which combined several different families of machine learning algorithm into a single decision making architecture were termed Ensemble. Though Naive Bayes and Bayesian Networks share common foundations, a decision was made to keep them separate since the latter includes the ability to incorporate domain knowledge to a larger degree, which was a motivating factor for its usage in several studies. A histogram depicting the usage patterns of the above algorithms, sorted according to frequency of use, can be seen in Figure \ref{algorithm_usage}.

\paragraph{Accuracy Reporting}  This review considered the accuracy measure defined as the proportion of the sum of total true positive and true negative classifications divided by all samples in the dataset, as the primary metric for analysis, thus ensuring that comparisons and interpretations were meaningful across different studies. 

Note that the charts will be presented in section \ref{lit-review} only include studies that reported accuracy as their measure of performance, however all results (including one paper that reported balanced accuracy, and three that reported the average ranked probability score) are presented in the summary tables. Moreover, results from binary-class and multi-class problems are not directly comparable since a given level of accuracy accuracy becomes harder to achieve as the number of classes increases. Thus, where a three-class formulation was used instead of a two-class formulation, these accuracies have also been excluded from the charts for comparative purposes, but again, are reported in the summary tables.

\paragraph{Future Research Themes} In order to draw insights across all papers as to what the general future research direction trends might be, a set of recurring general themes were first extracted from all the papers. Subsequently, all the text referring to future research directions across all the papers was coded based on the extracted themes, and a histogram was rendered depicting the frequency of each theme. Below is the list of themes used for this analysis:

\begin{itemize}
    \item improve machine learning parameterization
    \item improve training methods
    \item engineer richer features
    \item improve feature selection
    \item benchmark predictions with experts
    \item investigate the causal relationships deeper
    \item increase dataset size
    \item apply findings to a different sport
    \item use other machine learning algorithms
\end{itemize}

\section{Literature Review}
\label{lit-review}

In this section we review the literature in the field of ML for sport result prediction, in chronological order. We also summarize the results of the literature in tables in terms of the competition/league, the models used, the number of matches available in their original dataset, as well as the best performing model with the number of features used in that best model. The tables presented are: American Football - table \ref{tab:american_football_tbl}, Basketball - table \ref{tab:basketball-tbl}, Soccer - table \ref{tab:soccer-tbl}, Ice Hockey - table \ref{tab:ice-hockey-tbl}, Rugby Union - table \ref{tab:rugby-tbl} and other sports - table \ref{tab:other-sports-tbl}, which only had one study associated with them.

\begin{table}[tb]
\centering
\resizebox{\textwidth}{!}{%
\begin{tabular}{c P{4.5cm} P{3cm} P{2cm} P{2cm} P{2cm}}
\toprule
\toprule
\textbf{Paper} & \textbf{Models used} & \textbf{No. of features} & \textbf{No. of Matches} & \textbf{Accuracy of best model} \\
\midrule
\midrule
\cite{purucker1996neural} & ANN with BP*, Unsupervised: Hamming, ART, SOM & 6 & 90 & 75\% (week 14 and 15 combined) \\ \midrule
\cite{kahn2003neural} & ANN with BP & 10 & 208 & 75\% \\ \midrule
\cite{david2011nfl} & Committees of ANNs trained by LM & 11 & unknown & unknown \\ \midrule
\cite{delen2012comparative} & SVM, CART*, ANN & 28 & 244 & 86\% \\
\bottomrule
\bottomrule
\end{tabular}%
}
\caption{\textbf{American Football Studies (* denotes the best performing model).}}
\label{tab:american_football_tbl}
\end{table}

\subsection{1990s}
\cite{purucker1996neural} used an ANN as well as unsupervised learning techniques to predict the results of US National Football League (NFL) football matches in 1994, using data from weeks 11 to 16 of the competition (90 matches). Six features were considered: victories, yardage differential, rushing yardage differential, turnover margin, time in possession, and betting odds (the inclusion of betting line odds were found to improve upon initial results). An ANN trained with BP was used (BP provided the best performance among the different network training methods). Unsupervised learning methods were also applied, in particular: the Hamming, Adaptive Resonance Theory (ART), and Self Organizing Map (SOM) methods. The SOM provided the best performance among the unsupervised methods, but it generally could not match the performance of the ANN. Matches from weeks 12 to 15 were used to predict week 16, with the ANN correctly predicting 11 out of the 14 matches (78.6\%) in week 16. Weeks 12 to 14 were also used to predict week 15, with the ANN correctly predicting 10 of the 14 games (71.4\%). The author recognised that the dataset consisted of a small number of matches and features, and stated that improvements could be gained through fine tuning the encoding, the ANN architecture, and training methods.

\begin{table}[tb]
\centering
\resizebox{\textwidth}{!}{%
\centering
\begin{tabular}{c P{2.5cm} P{4cm} P{2cm} P{2cm} P{2cm}}
\toprule
\toprule
\textbf{Paper} & \textbf{Competition} & \textbf{Models used} & \textbf{No. of features} & \textbf{No. of Matches} & \textbf{Accuracy of best model} \\ 
\midrule
\midrule
\cite{loeffelholz2009predicting} & NBA & ANN (types: FFNN*, RBG, PNN, GRNN, fusions of these) & 4 & 650 & 74.3\% \\ \midrule
\cite{zdravevski2009system} & NBA & All models in WEKA (Logistic Regression*) & 10 & 1230 & 72.8\% \\ \midrule
\cite{ivankovic2010analysis} & Serbian First B & ANN trained with BP & 51 & 890 & 81\% \\ \midrule
\cite{miljkovic2010use} & NBA & kNN, Decision Tree, SVM, Naïve Bayes* & 32 & 778 & 67\% \\ \midrule
\cite{cao2012sports} & NBA & Simple Logistic Regression*, Naïve Bayes, SVM, ANN & 46 & 4000 & 67.8\% \\ \midrule
\cite{shi2013predicting} & NCAAB & ANN*, C4.5 Decision Tree, RIPPER, Random Forest & 7 & 32236 & 74\% \\ \midrule
\cite{thabtah2019nba} & NBA & ANN, Naïve Bayes, LMT Decision Tree* & 8 & 430 & 83\% \\ \bottomrule
\bottomrule
\end{tabular}%
}
\caption{\textbf{Basketball Studies (*denotes the best performing model).}}
\label{tab:basketball-tbl}
\end{table}

\subsection{2000s}
\cite{kahn2003neural} predicted matches in the NFL using data from 208 matches in the 2003 season. An ANN with BP was used, and the features included in the model were: total yardage differential, rushing yardage differential, time in possession differential, turnover differential, a home or away indicator, home team outcome and away team outcome. The numeric features were calculated as the 3-week historical average (the average of the statistic over the past 3 weeks of the competition), as well as the season average (the average of the statistic over the entire season). It was found that using the season average provided better accuracy. Weeks 1 to 13 of the 2003 competition were used as training data, with weeks 14 and 15 as the test set. Accuracy of 75\% was achieved, which was slightly better than expert predictions on the same games. It was suggested that in future work, betting odds or team rankings could be included as model predictors, and also that previous seasons could be used for training a model instead of only the current season.

\begin{table}[tb]
\centering
\resizebox{\textwidth}{!}{%
\begin{tabular}{c P{2.5cm} P{5cm} P{2cm} P{2cm} P{2cm}}
\toprule
\toprule
\textbf{Paper} & \textbf{Competition} & \textbf{Models used } & \textbf{No. of features} & \textbf{No. of Matches} & \textbf{Accuracy of best model} \\ 
\midrule
\midrule
\cite{weissbock2013use} & NHL & Naive Bayes, SVM, ANN*, C4.5 Decision Tree & 11 & 517 & 59\% \\ \midrule
\cite{weissbock2014combining} & NHL & Weka: ANN, Naive Bayes, Complement Naive Bayes, Multinomial Naive Bayes, LibSVM, SMO, J48, JRip, Logistic, Simple Logistic, Simple Naive Bayes, Cascading Ensemble* & 6 & 708 & 60.3\% \\ \midrule
\cite{gu2019game} & NHL & KNN, SVM, Naive Bayes, Discriminant Analysis, Decision Trees + ensembles of these: Boosting*, Bagging, AdaBoost, RobustBoost & 19 & 1230 & 91.8\% \\
\bottomrule
\bottomrule
\end{tabular}%
}
\caption{\textbf{Ice Hockey Studies (*denotes the best performing model).}}
\label{tab:ice-hockey-tbl}
\end{table}

\cite{o2004evaluation} were interested in comparing human experts with computer-based methods in terms of their ability to predict the results of matches in the 2003 Rugby Union\footnote{Rugby Union will hereafter be referred to simply as Rugby (not to be confused with Rugby League, which is a different sport).} World Cup. Multiple Linear Regression, Logistic Regression, an ANN and a Simulation Model were the computer-based models. The authors made use of data from the previous four World Cups to develop predictive models of match outcomes based on three variables: team strength (determined by synthesising world rankings\footnote{World rankings in Rugby had not yet been introduced at the time of the 2003 Rugby World Cup.}), distance travelled to the tournament as a measure of home advantage, and the number of recovery days between matches. The computer-based models predicted between 39 and 44.5 of the 48 matches correctly, while the 42 human experts correctly predicted an average of 40.7 matches. However, it was noted that there was a far greater variation in the accuracies from the human experts, with the most successful person correctly predicting 46 of the 48 matches. The most successful computer-based model was the Simulation Model.

\cite{reed2005development} built seven predictive models (including models based on Multiple Linear Regression, an ANN and Discriminant Analysis) using seven predictor variables, to predict results in English Premier League (EPL) Soccer and Premiership Rugby in England. The dataset was derived from three football teams and two Rugby teams, across three seasons. After a review, the authors included the following predictive variables: match venue, rest, the team's position on league table, the opposition team's position on league table, distance travelled to the match (for both teams), and form. These variables were common to both Rugby and Soccer, i.e., the study did not incorporate variables specific to in-play match events in Rugby or Soccer. Considering that a draw is a much more common outcome in Soccer than in Rugby, it was surprising that the percentage of Soccer matches (57.9\%) that were predicted correctly was higher than Rugby (46.1\%). Despite the best performing predictive model accuracies being rather low, the models still outperformed the predictions of human experts with an ANN found to yield the best accuracy. The authors stated that motivational, injury and other variables were not included, and that future studies could concentrate on developing complex pattern recognition technology rather than using simple linear models which could lack flexibility.

\cite{joseph2006predicting} found that it may be beneficial to incorporate expert knowledge into a modeling process. Expert knowledge was incorporated while constructing a Bayesian Network (BN) model, and it was found that including such knowledge can result in strong performance, especially when the sample size is small (which is often the case with sports). A decision tree (MC4) and kNN model were also used to predict the results of Soccer matches played by the EPL team Tottenham. Their dataset consisted of 76 matches. Four variables were included in the expert model, while 30 variables were used in their original general model. The expert BN was found to provide the best performance, achieving 59.2\% accuracy in predicting a home win, away win or a draw (a 3-class problem). In the future, the authors proposed developing a more symmetrical model using similar data but for all the teams in the EPL, and also incorporating player quality features (e.g., players that have performed at international level) into the model. In addition, they mentioned that additional nodes could be added to the BN, e.g., attack quality and defence quality.

\cite{mccabe2008artificial} considered four different sports: Rugby League in Australia's National Football League (NFL), Australian Rules Football in the Australian Football League (AFL), Super Rugby, and EPL soccer, using data from 2002 to 2007. An ANN trained with BP and an ANN trained with Conjugative-Gradient Descent (CGD) were applied. It was found that training the ANN with BP was slightly more accurate than the CGD algorithm; however, the training time with BP was longer. The variables used in the study were the same across all of the four sports, i.e., sport-specific variables derived from in-play events within a match were not included. The average accuracy achieved by the ANN with BP was 67.5\%, higher than expert predictions which ranged from 60\% to 65\%. For future work, the authors mentioned that other sports could be considered, as well as including more variables, and predicting the points margin.

\cite{loeffelholz2009predicting} attempted to predict National Basketball Association (NBA) matches from the 2007/2008 season. There were 620 matches that were used for training and testing, and 30 that were used as the validation set - treated as “un-played” games. The set of predictive variables consisted of: field goal percentage, three-point percentage, free-throw percentage, offensive rebounds, defensive rebounds, assists, steals, blocks, turnovers, personal fouls, and points. In terms of predicting the un-played games, averages of variables in the current season were found to result in better performance than averaging statistics across the past five matches. Fusion of ANNs was an approach also investigated by the authors, using Bayesian Belief Networks and Neural Network (NN) fusion in particular. Four different types of ANN were applied - a Feed-Forward NN (FFNN), a Radial Basis Function NN (RBF-NN), a Probabilistic NN (PNN) and a Generalized NN (GRNN). Their best model - an ANN with FFNN with four shooting variables, correctly predicted the winning team 74.3\% of the time on average, which was better than experts from USA Today 2008 who achieved 68.7\%. The feature selection technique used was an iterative Signal-to-Noise Ratio (SNR) method \citep{bauer2000feature}, which selected four out of the 22 original variables. It was mentioned that although fusion techniques did not result in higher accuracy on this occasion, they still could be worth investigating in the future. Also, different features could be used in the baseline model, and the models created could be adjusted to determine if the model can beat the betting odds rather than simply predicting which team will win.

\cite{zdravevski2009system} obtained two seasons of NBA data consisting of 1230 matches. The first season was used as the training dataset, while the second was used as the test dataset. All of the algorithms contained in the WEKA machine learning workbench were used for experimentation, applied with their default parameter settings. A set of 10 variables was selected by the authors. Classification accuracy of 72.8\% was achieved with Logistic Regression. It was stated that in future work it would be preferable to compare their predictions to those of experts, and that it might be possible to cluster training and test data to use different models on each cluster in order to account for winning and losing streaks. Furthermore, it was mentioned that aggregations or ensembles of classifiers (e.g., voting) could be investigated, and that automatic feature selection methods should be used rather than manual human selection.

\subsection{2010s}
\subsubsection{2010 - 2014} \hfill\\

\cite{buursma2010predicting} used data from 15 years of Dutch Soccer to predict results. Buursma was specifically interested in: which variables were important when predicting match outcomes, how the probabilities for the match outcomes can be predicted from these variables, and what bets should be placed to maximize profit. The author experimented with the following models implemented in WEKA: Classification via Regression, a Multi-class Classifier, Rotation Forest, Logit Boost, Bayesian Network, Naive Bayes, and ZeroR. There were three match outcomes to be predicted: home win, draw, and away win. The feature set consisted of 11 features, and all features were either aggregated or averaged across a team’s past 20 matches (20 was found by experimentation to be the best number of matches in which to average across). It was determined that Classification via Regression and the Multi-class Classifier had the best prediction accuracy of 55\%. For future work the author considered including more features such as: yellow/red cards, the number of players each team has, their managers, their player budget and their home ground capacity. Buursma was also interested in applying such a model to other similar sports such as basketball, baseball and ice hockey in the future.

\cite{huang2010neural} used an ANN with BP to predict match results in the 2006 Soccer World Cup. A total of 64 matches were played in the tournament. The output of the ANN was the relative ratio between two teams for each game, which could then be converted into an odds ratio. Accuracy of 76.9\% accuracy was achieved; however, it should be noted that this is higher than some other Soccer studies because their model did not include prediction of draws. Feature selection was performed based on domain knowledge of the authors. Eight features were used in the model, obtained from the match statistics of the FIFA World Cup website. These were: goals for, shots, shots on goal, corner kicks, direct free kicks on goal, indirect free kicks on goal, possession, and fouls conceded. The authors remarked that the ANN generally had difficulty predicting a draw, which was a frequent outcome in the group stages of the tournament.

\cite{ivankovic2010analysis} utilized an ANN with BP for predicting the results of the Serbian First B Basketball League using five seasons of data from 2005/2006 to 2009/2010 - comprising 890 matches. The authors applied the Cross-Industry Standard Process for Data Mining (CRISP-DM) framework \citep{wirth2000crisp} as their experimental approach, and investigated how the successful shot percentage in six different regions of the court affected match results. The input dataset was divided into 75:25 ratios for training and testing, and 66.4\% accuracy was obtained on the test set. The authors then cycled back to the data preparation phase of the CRISP-DM framework to see if adding additional variables could improve results. In particular, offensive rebounds, defensive rebounds, assists, steals, turnovers and blocks were subsequently incorporated. This improved the accuracy achieved by the model to just under 81\%. It was concluded that actions in the zone directly under the hoop - rebounds in defence, as well as scoring in this zone were crucial in determining the outcome of the game. It was mentioned that a richer data set and new software solutions may be helpful in the future, so that all relevant events are included.

\begin{table}[tb]
\centering
\resizebox{\textwidth}{!}{%

\begin{tabular}{c P{2.5cm} P{4cm} P{2cm} P{2cm} P{2cm}}
\toprule
\toprule
\textbf{Paper} & \textbf{Competition} & \textbf{Models used} & \textbf{No. of features} & \textbf{No. of Matches} & \textbf{Accuracy of best model} \\ \hline

\midrule
\cite{o2004evaluation} & 2003 Rugby World Cup & Multiple linear regression, Logistic Regression, ANN, Simulation model* & 3 & 48 & 93\% \\ \midrule
\cite{reed2005development} & English Premiership Rugby & Multiple linear regression, ANN*, discriminant function analysis & 7 & 498 & 46.1\% \\ \midrule
\cite{mccabe2008artificial} & Super Rugby & ANN trained with BP* and CGD & 19 & unknown & 74.3\% \\ \midrule
\cite{o2016predictive} & 2015 Rugby World Cup & Linear regression*, Simulation & 1 & 280 & 74.4\% \\
\bottomrule
\bottomrule
\end{tabular}%
}
\caption{\textbf{Rugby Union Studies (*denotes the best performing model).}}
\label{tab:rugby-tbl}
\end{table}

\cite{miljkovic2010use} used data from 778 games in the regular 2009/2010 NBA season to predict basketball match outcomes. The predictive features were divided into game statistics which directly relate to events within the match e.g., fouls per game and turnovers per game, and those that relate to standings e.g., total wins and winning streak. Naive Bayes was found to be the best performing model compared to kNN, a Decision Tree and a Support Vector Machine (SVM) model. Accuracy of 67\% was achieved with Naive Bayes, using 10-fold cross validation. Their future research plans included applying their system to other sports, and to experiment with other models such as ANNs.

\cite{david2011nfl} used a committees-of-committees approach with ANNs to predict NFL matches, whereby many networks were trained against different random partitions of the data. Under their approach, 500 ANNs were used for training, and from this, the best 100 were used in each committee. Subsequently, 50 such committees were used to produce the final predictive results. The mean was used to determine the vote of each committee and to combine the predictions of the committees. The features used were differentials between home and away teams based on: passing yards, yards per rush play, points, interceptions, and fumbles. The differentials between home and away teams were to incorporate the well-known home advantage phenomenon. The season-to-date average of these statistics were used describe each team, apart from the first five rounds of the season where the weighted average between the current season and the previous season statistics was used\footnote{100\% of the statistics from the previous season were used for week 1, then 80\% of the previous season and 20\% of the current season in week 2. This process was continued with a 20\% increase each week until week 6, at which point only the current season statistics were used.}. A total of 11 inputs were used in the ANN for each game, and the Levenberg Marquadt (LM) routine was used to train the network. Principal Components Analysis (PCA) and derivative-based analyses were applied in order to determine which variables were most influential. It was found that 99.7\% of the variance in their data was due to passing and rushing differentials. The results were compared to bookmakers, as well as predictions from the website thepredictiontracker.com. It was found that the results were comparable to the predictions of bookmakers, and were also better than most of the predictive models on the website. The avenues for future experimentation were listed as: using different types of ANN, e.g., RBF, including additional statistics (e.g., possession, strength-of-schedule, kicking game and injuries), investigating how to best predict games early in the season, and application to other levels of football (e.g., NCAA) as well as to other sports.

In contrast to \cite{joseph2006predicting}, \cite{hucaljuk2011predicting} found that incorporating expert opinion did not produce any improvement in predicting the outcome of soccer matches. Their dataset consisted of 96 matches (6 rounds) in the regular part of the European Champions League competition. This data was split three different ways into training and test datasets: a 3 round-3 round training-test split, a 4 round-2 round split, and a 5 round-1 round split. Feature selection resulted in 20 features in their basic feature set. An additional feature set consisted of the 20 basic features plus variables selected by experts. Naive Bayes, Bayesian Networks, Logit Boost, kNN, a Random Forest and an ANN with BP were used in the experiments. Overall, the ANN performed the best, achieving classification accuracy of 68\%. Perhaps surprisingly, the expert-selected features were not found to yield any improvement. It was mentioned that further improvements could be gained in refining feature selection, modeling the form of players in matches throughout a season, as well as obtaining a larger dataset.

\begin{table}[htb]
\captionsetup{font=scriptsize}
\centering
\resizebox{\textwidth}{!}{%
\begin{tabular}{c P{3cm} P{5cm} P{2cm} P{2cm} P{2cm}}
\toprule
\toprule
\textbf{Paper} & \textbf{Competition} & \textbf{Models used} & \textbf{No. of features} & \textbf{No. of Matches} & \textbf{Accuracy of best model} \\
\midrule
\midrule

\cite{reed2005development} & English Premier League & Multiple Linear Regression, ANN*, Discriminant Function Analysis & 7 & 498 & 57.9\% (3-class) \\ \midrule
\cite{joseph2006predicting} & English Premier League - Tottenham Hotspur & Bayesian Network, Expert Bayesian Network*, Decision Tree, kNN & 4 & 76 & 59.2\% (3-class) \\ \midrule
\cite{mccabe2008artificial} & English Premier League & ANN trained with BP* and CGD & 19 & unknown & 54.6\% (3-class) \\ \midrule
\cite{buursma2010predicting} & Dutch Eredivisie League & WEKA: MultiClassClassifier with ClassificationViaRegression*, RotationForest, LogitBoost, Bayesian Network, Naïve Bayes, ZeroR & 11 & 4590 & 55\% (3-class) \\ \midrule
\cite{huang2010neural} & 2006 Soccer World Cup & ANN trained with BP & 8 & 64 & 62.5\% (3-class), 76.9\% (2-class)\\ \midrule
\cite{hucaljuk2011predicting} & European Champions League & naïve Bayes, Baysian network, LogitBoost, kNN, random forest, ANN* & 20 & 96 & 68\% (3-class) \\ \midrule
\cite{odachowski2012using} & Various leagues & BayesNet*, SVM, LWL, Ensemble Selection, CART, Decision Table & 320 & 1,116 & 70.3\% (2-class), 46\% (3-class) \\ \midrule
\cite{tax2015predicting} & Dutch Eredivisie League & WEKA: NaiveBayes, LogitBoost*, ANN, RandomForest, CHIRP, FURIA, DTNB, J48,HyperPipes & 5 & 4284 & 56.1\% (3-class) \\ \midrule
\cite{prasetio2016predicting} & English Premier League & Logistic Regression & 4 & 2280 & 69.5\% (2-class)\\ \midrule
\cite{danisik2018football} & Various leagues & LSTM NN classification, LSTM NN regression*, Dense Model & 139 & 1520 & 52.5\% (3-class), 70.2\% (2-class)\\ \midrule
\cite{hubavcek2019learning} & 52 leagues & XGBoost classification*, XGBoost regression, RDN-Boost & 66 & 216,743 & 52.4\% (3-class) \\ \midrule
\cite{constantinou2019dolores} & 52 leagues & Hybrid Bayesian Network & 4 & 216,743 & 51.5\% (3-class) \\ \midrule
\cite{berrar2019incorporating} & 52 leagues & XGBoost*, kNN & 8 & 216,743 & 51.9\%** (3-class) \\
\bottomrule
\bottomrule
\end{tabular}%
}
\caption{\textbf{Soccer Studies (*denotes the best performing model).} Accuracies for 2-class (winn,loss) and 3-class (win,loss,draw) problems are denoted. **Berrar (2019)'s best performing model was kNN, but post-competition they mention that they improved on this with XGBoost. The accuracy of 51.9\% is for their in-competition result - the XGBoost accuracy was not reported in their paper but it would have been slightly higher than this. }
\label{tab:soccer-tbl}
\end{table}
\afterpage{\clearpage}

\cite{cao2012sports} built a system that automated data collection and applied various data mining algorithms for basketball match prediction. Six years of data from NBA matches was obtained, from the 2005/2006 season to the 2010/2011 season (comprising around 4,000 matches). The dataset was divided into training sets, test sets and validation sets. Four classification algorithms were used in the experiments: Simple Logistic Regression, Naive Bayes, a SVM and an ANN with BP. The feature selection process was manual, with 46 features being selected based on the domain knowledge of the author. All models were found to produce reasonably similar rates of classification accuracy, with simple Logistic Regression achieving the best accuracy of 67.8\%. The Simple Logistic Regression model also provided automatic feature selection. The best expert predictions on teamrankings.com were slightly better, achieving 69.6\% accuracy. The author suggested that in future clustering techniques could be used to group players by positional group, or to identify outstanding players. Using outlier detection methods to identify outstanding players or team status, investigating the impact of player performance match outcome, and comparing different feature sets derived from box-score and team statistics were also mentioned as avenues for future work.

\cite{delen2012comparative} used a SVM, CART Decision Tree and an ANN to predict the outcome of NCAA Bowl American Football matches. The CRISP-DM framework was applied as the experimental approach. The dataset was made up of 28 features and 244 matches. Both a classification approach predicting home win/away win, as well as a numeric prediction approach, which predicted the points margin (home team points less away team points), were compared. The CART Decision Tree provided the best performance with 86\% accuracy (using 10-fold CV), which was statistically significantly better than the other models. The classification approach was found to produce better results compared with numeric prediction. When models were trained on the seasons 2002/2003 to 2009/2010, and then tested on the 2010-2011 season, CART again had the best performance with 82.9\% accuracy. The suggested directions for future research were: including more variables or representing them in different forms, making use of other classification and regression methods (e.g., rough sets, genetic algorithm based classifiers, ensemble models), experimenting with seasonal game predictions by combining static and time series variables, and applying the approach to other sports.

\cite{odachowski2012using} analysed fluctuations in betting odds and used ML techniques to investigate the value in using such data for predicting soccer matches. The odds for home win, away win and draw for the preceding 10 hours, measured at 10-minute intervals, were tracked over time (the data having been obtained from the Betfair and Pinnacle Sports websites). A total of 32 features were computed from this time-series - e.g., maximum and minimum changes in betting odds, overall changes in odds, and standard deviations. The 10 hour period was divided into thirds and the features were also calculated based on these sampling periods. The authors balanced their dataset such that there were an equal number of home wins, away wins, and draws (372 matches for each class). Six classification algorithms from WEKA were compared: Bayesian Network, SMO, LWL, ensemble selection, Simple CART and a Decision Table, and feature selection methods contained in WEKA were applied. It was found that draws in particular were difficult to correctly predict, with only around 46\% accuracy obtained when attempting 3-class classification. However, accuracy of around 70\% was obtained when ignoring draws. Discretization and feature selection methods were found to improve results. The authors suggested that additional features describing changes in betting odds could be included going forward. 

\cite{shi2013predicting} investigated the viability of machine learning in the context of predicting outcomes of individual NCAAB matches which up to this point had been dominated by statistical methods. They used the WEKA toolkit for their comparative study of accuracies between ANN, C4.5, RIPPER and Random Forest algorithms. Experiments were conducted using data from six seasons together with an expanding window approach so that initially training was performed on the 2008 season and the testing on the 2009 season. Thereafter, the combination of all previous seasons composed the training set until the 2013 season. The authors concluded that on average, the ANN algorithm with default parameter settings provided the best accuracies, though statistical tests to confirm this were not provided. The top ranked features in terms of importance were location, the so-called "four factors" \citep{oliver2002basketball}, and adjusted offensive and defensive efficiencies (\url{kenpom.com}). The authors also remarked that they experienced an upper limit of 74\% accuracy which they could not improve beyond, while noting that feature engineering and selection hold promise for an improvement in results.  

\cite{weissbock2013use} noted that Ice Hockey had not received much attention in ML research with respect to predicting match outcomes. The continuous nature of this sport makes it difficult to analyse due to a paucity of quantifiable events such as goals. This characteristic was cited as a possible reason for the lack of attention Hockey had received historically. \cite{weissbock2013use} focused on exploring the role of different types of features in predictive accuracies across several types of ML algorithms. They considered both traditional statistics as features, as well as performance metrics used by bloggers and statisticians employed by teams. They used WEKA's implementations of ANN, Naive Bayes, SVM and C4.5 for training classifiers on datasets describing National Hockey League (NHL) match outcomes in the 2012/2013 season. The entire dataset amounted to 517 games. The authors concluded that traditional statistics outperformed the newer performance metrics in predicting outcomes of single games using 10-fold CV, while the ANN displayed the best accuracy results of 59\%. Research in extracting more informative features and predicting winners of the NHL playoffs was cited as future work as well as incorporating knowledge from similar sports such as soccer.

\cite{weissbock2014combining} combined statistical features with features derived from pre-game reports, in order to determine whether the sentiment of the pre-game reports had any usefulness for the prediction of NHL Ice Hockey matches. Data from 708 NHL games in the 2012/2013 season was collected, and pre-game reports were acquired from the website NHL.com. Both natural language processing and sentiment analysis based features were used in the experiments. The three statistical features that were used were identified from their previous work \citep{weissbock2013use}, which were: cumulative goals against and their differential, and the match location (home/away). The following algorithms from the WEKA toolkit were applied with default parameters: ANN, Naive Bayes, Complement Naive Bayes, Multinomial Naive Bayes, LibSVM, SMO, J48 Decision Tree, JRip, Logistic Regression and Simple Logistic Regression. Three models were compared: models that used statistical features only, models using pre-game report text only, and models trained with sentiment analysis features. It was found that using only pre-game reports features for prediction did not perform as well as models trained with statistical features only. A meta-classifier with majority voting was implemented where the confidence and predicted output from initial classifiers was fed into a second layer. This architecture provided the best accuracy of 60.25\%. This cascading ensemble approach on the statistical feature set provided superior performance to using both feature sets, suggesting that the pre-game reports and statistical features provided somewhat different perspectives. The authors noted that it was difficult to predict matches with a model trained beyond one or two seasons due to player changes and so on.

\subsubsection{2015 - 2019} \hfill\\

\cite{tax2015predicting} focused on identifying features which have a high utility for predicting football match outcomes, and which satisfy the criterion of being easily retrievable. In particular, they sought to explore whether or not there were different levels of utility between features broadly classified as public match-data sources and those derived from bookmaker odds. Their experiments covered data from the 2000 to 2013 seasons of the Dutch Eredivisie competition. The researchers conducted numerous experiments comparing the Naive Bayes, LogitBoost, ANN, Random Forest, CHIRP, FURIA, DTNB, J48 and Hyper Pipes algorithms from the WEKA toolkit. The experiments were performed in conjunction with feature reduction methods such as PCA, Sequential Forward Selection and ReliefF. The best results on the public match-data sources were achieved by the Naive Bayes and ANN classifiers in combination with PCA, achieving an accuracy of 54.7\%. FURIA achieved the highest accuracy using bookmaker odds features with 55.3\%; however, this was ultimately not found to be statistically significant. A marginal improvement in accuracy was realised with LogitBoost and ReliefF when both bookmaker odds and public match-data features were used, producing an accuracy of 56.1\%; however, this also was not at a statistically significant level, but nonetheless pointed to the potential utility in combining a broader variety of features for further investigation.

\begin{table}[tb]
\centering
\resizebox{\textwidth}{!}{%
\begin{tabular}{c P{2.5cm} P{4cm} P{2cm} P{2cm} P{2cm}}
\toprule
\toprule
\textbf{Paper} & \textbf{Sport-Competition} & \textbf{Models used} & \textbf{No. of features} & \textbf{No. of Matches} & \textbf{Accuracy of best model} \\
\midrule
\midrule
\cite{mccabe2008artificial} & Rugby League - NRL & ANN trained with BP* and CGD & 19 & unknown & 63.2\% \\ \midrule
\cite{mccabe2008artificial} & Australian Rules Football - AFL & ANN trained with BP* and CGD & 19 & unknown & 65.1\% \\ \midrule
\cite{pathak2016applications} & Cricket - International ODIs & Naïve Bayes, Random Forest, SVM* & 4 & unknown & 61.7\% (balanced accuracy) \\ \midrule
\cite{valero2016predicting} & Baseball - MLB & Lazy Learners, ANN, SVM*, Decision Tree & 3 & 24,300 & 59\% \\
\bottomrule
\bottomrule
\end{tabular}%
}
\caption{\textbf{Other Sport Studies (*denotes the best performing model).}}
\label{tab:other-sports-tbl}
\end{table}

\cite{pathak2016applications} attempted to predict the results of One-Day International (ODI) Cricket. The authors included four predictive variables based on the prior statistical work of \cite{bandulasiri2008predicting}: toss outcome, whether the game was played at home or away, whether the game was played during the day or at night, and whether the particular team in question batted first or second. Three classification models were applied - Naive Bayes, Random Forest and SVM. Data for matches from 2001 - 2015 was collected from cricinfo.com. A separate model was constructed for each team, which analysed that team with respect to all other teams. An 80-20 training-test split was used. To mitigate the effect of imbalanced data sets (some teams had a high ratio of wins to losses), the three models were evaluated based on balanced accuracy and the Kappa statistic. SVM was found to perform the best across all teams with an average balanced accuracy of 61.67\%. It was suggested that using more novel methods of classification for outcome prediction, including additional features, application to other forms of cricket (e.g. test matches, T20), and application to other sports (e.g. baseball, football) could be performed in future work.

\cite{prasetio2016predicting} used Logistic Regression to predict EPL soccer results in the 2015/2016 season. Their data set consisted of data from six seasons from the 2010/2011 to the 2015/2016 seasons - a total of 2,280 matches. Home offense, away offense, home defence and away defence were used as the input features, and it was found that the significant variables were home defence and away defence (it was not mentioned in the paper how these offense and defence ratings were actually constructed). Despite this, the model that included all four variables was found to yield a higher accuracy. The split of training and testing data was varied into four different training-test partitions which produced four different sets of model coefficients. The best performing model achieved 69.5\% accuracy (however since they used a binary Logistic Regression model, this result was based on the omission of draws). In the future, they remarked that the results could be used to assist management with game strategy, or that the trained models could be turned into a recommendation system for evaluating which players to purchase.

\cite{valero2016predicting} performed a comparative study for prediction of Baseball matches, using 10 years of Major League Baseball (MLB) data. Lazy Learners, ANNs, SVM and Decision Trees were the candidate models. Inspired by \cite{delen2012comparative}, the CRISP-DM framework was applied, and additionally, the performance of a classification approach (win or loss for the home team) as well as a numeric prediction approach (predicting the run difference between the home and away team) were compared. Feature selection methods from WEKA were applied for dimensionality reduction and ranking of the original set of 60 features. Their model used only the top three ranked variables - home field advantage, Log5 and the Pythagorean Expectation\footnote{The Pythagorean Expectation was developed by the Baseball statistician Bill James \citep{james1984bill}, and represents the expected number of wins for a team given their runs scored and runs allowed, while Log5 ratings are essentially the same as \cite{elo1978rating} Ratings \citep{koseler2017machine}.} (the addition of more variables was not found to improve results). SVM was found to produce the best accuracy under both the classification and numeric prediction approaches. Consistent with \cite{delen2012comparative}, the classification approach performed significantly better than the numeric prediction approach. The SVM classification model achieved accuracy of around 59\%. However, when using the 2005-2013 seasons as training data and the 2014 season as test data, the model's predictions were not significantly more accurate than predictions derived from match betting odds. The authors highlighted the difficulty in predicting outcomes in Baseball using statistical data alone, but suggested that experiments using the Japanese or Korean Baseball leagues could be useful. In the future, the authors also considered adjusting their model parameters, refining features and extending their datasets. In addition, they planed to apply the model to other sports such as Basketball, Football and Water Polo. 

\cite{o2016predictive} compared the accuracy of 12 models based on Linear Regression to analyse the effectiveness of predicting team performances in the 2015 Rugby World Cup. The models differed in (i) whether or not the assumptions of the predictive modeling technique were satisfied or violated, (ii) whether all (1987-2011) or only recent Rugby World Cup tournaments (2003-2011) data were used, (iii) whether the models combined pool and knockout stage match data, and (iv) whether the models included a variable that tried to capture a relative home advantage. The common independent variable in all models was the relative quality, which represented the difference between the higher ranked team’s world ranking points. The dependent variable was the point difference in the match outcome between two teams. All models were executed 10,000 times within a simulation package that introduced random variability. The best model achieved accuracy of 74\%, with the key finding being that match outcomes in international Rugby appeared more difficult to predict than in previous years. The best model used data from all previous Rugby World Cups in a way that violated the assumptions of Linear Regression. The best model used only one independent variable and ignored the relative home advantage, while generating separate models for the pool and knockout stage matches.

\cite{danisik2018football} applied a Long-Short-Term-Memory (LSTM) NN model \citep{hochreiter1997long} for match result prediction in a number of different Soccer leagues. Classification, numeric prediction and dense approaches were compared, and were also contrasted with an average random guess, bookmaker's predictions, and the most common class outcome (home win). Player-level data was included, and was obtained from FIFA video games. Incorporating player-level and match history data, a total of 139 features were included (134 in the dense model). Four seasons of the EPL were considered - 2011/2012 to 2015/2016, comprising a total of 1,520 matches.
The average prediction accuracy obtained for a three-class classification problem was 52.5\%, achieved with the LSTM Regression model. This result was achieved using 2011/2012, 2012/2013 and 2015/2016 seasons as the training dataset and the 2013/2014 season as the validation dataset. The accuracy obtained for a two-class problem, excluding draws, was 70.2\%. It was stated that betting odds and some additional match-specific, player and team features could be included in the future. Additionally, the use of convolution to transform input attributes during training could be investigated as well as a deeper exploration the the ability of LSTM NN to leverage features like specific tactics.

\cite{thabtah2019nba} used Naive Bayes, an ANN, and a LMT Decision Tree model to forecast the outcome of NBA Basketball matches. They focused on testing different feature sets in order to find the optimal subset of predictors. The dataset was obtained from Kaggle.com, and contained 430 NBA finals matches from 1980 to 2017 and 21 features. The target variable was a binary variable expressed as a win or a loss. It was found that defensive rebounds was the most important factor influencing match results. The feature selection methods used were based on Multiple Regression, Correlation Feature Subset - CFS \citep{hall1998correlation}, and RIPPER \citep{cohen1995fast}. Defensive rebounds were selected as important features by all three feature selection methods. The best performing model in terms of accuracy was trained on a feature set containing eight variables. These were selected from the RIPPER decision rules and subsequently trained using the LMT model, which generated 83\% accuracy. The authors suggested that larger datasets and more attributes could be considered going forward, such as the team's players and coach. The use of other models such as function based techniques and deep learning methods were also stated as potential avenues for further research.


\cite{gu2019game} reported that ensemble methods provided encouraging results in the prediction of outcomes of NFL Hockey matches over multiple seasons. The data extraction was automated and scraped from several websites listing NHL matches from 2007/2008 to 2016/2017 seasons (1230 matches). Data was merged from several sources including historical results, opposition information, player-level performance indicators, as well as player ranks using PCA. A total of 26 team performance variables were also included in their model. The kNN, SVM, Naive Bayes, Discriminant Analysis and Decision Tree algorithms, together with several ensemble-based methods, were applied. The ensemble-based methods (Boosting, Bagging, AdaBoost, RobustBoost) achieved the highest accuracy on the test set, with 91.8\%. In terms of future research, the authors mentioned that additional data could further improve predictions, features consisting of different or additional predictors could be used, e.g., player ranking metrics, moving-averaged or exponentially-smoothed team and player performance metrics, psychological factors, expert/coach strategic or tactical judgements and player's physical or mental assessments could be included. They also mentioned that the ML problem could be re-formulated so that a different outcome could be predicted, e.g., whether or not a team will make the playoffs or which team will win the championship, by training a model on the regular season data.

\paragraph{\textbf{The Open International Soccer Database: Prediction Challenge}}

In 2017, a significant development took place for ML researchers interested in Soccer. A comprehensive open-source database called the \textit{Open International Soccer Database} was compiled and made public. The database contains over 216,000 matches from 52 leagues and 35 countries \citep{dubitzky2019open}. The motivation behind this project was to ultimately encourage ML research in Soccer by building an up-to-date knowledge base that can be used on an ongoing basis for the prediction of real-world soccer match outcomes, as well as a benchmark dataset which makes comparisons between experiments more robust. In order to maximise the utility of this database, a deliberate design choice was made to collect and integrate only data that are easily available for most soccer leagues worldwide, including lower leagues. The consequence of this is that the database lacks fields that are highly-specialized and sophisticated. 

Subsequent to its creation, the 2017 Soccer Prediction Challenge was conducted, which hosted a competition based on this dataset. The results arising from it were published in a special issue of the Machine Learning (Springer) Journal. The challenge involved building a single model in order to predict 206 future match outcomes from 26 different Soccer leagues, which were to be played from March 31 to April 9 in 2017. Unlike most prior studies that used accuracy as a performance metric, this competition used the average Ranked Probability Score (RPS) (\cite{epstein1969scoring}, \cite{constantinou2012pi}). In the remainder of this section, we summarize three of the papers from the competition that focused on ML.

\cite{hubavcek2019learning} experimented with both relational and feature-based methods to learn predictive models from the database. Pi-ratings \citep{constantinou2013determining}, which capture both the current form and historical strengths of teams, and a rating based on PageRank \citep{page1999pagerank} were computed for each team. XGBoost gradient boosted tree algorithms (both regression and classification) were employed as the feature-based method, while RDN-Boost was used as the relational method. Their feature-based classification method with the gradient boosted tree algorithm performed best on both the validation set and the unseen challenge test set, with the model achieving 52.4\% accuracy on the test set. Possibilities for future work were cited as: augmenting the feature set, weighting aggregated data by recency, including expert guidance in forming relational concepts, e.g., using active learning, and identification of features that are conditionally important given occurrences of certain features at higher levels of the tree.

\cite{constantinou2019dolores} created a model that combined dynamic ratings with a Hybrid Bayesian Network. The rating system was based partly on the Pi-rating system of \cite{constantinou2013determining}, computing a rating that captures the strength of a team relative to others in a particular competition. Pi-ratings consider the actual goal differences in a match, compared with the expected result (based on existing ratings) in order to update the team ratings. The rating calculation involves placing more emphasis on the result rather than the margin, so the effect of large goal differences are dampened. Unlike the original Pi-ratings, this version also incorporated a team form factor. This ratings algorithm essentially searched for continuously for over or under-performance - see the paper for further details. Four ratings features (two each for the home and away team) were used as the inputs into the Bayesian Network. The model provided empirical evidence that a model can make good predictions for a match between two particular teams, even when the prediction was based on historical match data that involved neither of those two teams. The model provided accuracy of 51.5\% on the challenge test data set. The author recognized the limited nature of this data set, and that incorporating other key factors, such as player transfers, the availability of key players, participation in international competitions, a new coach, level of injuries, attack and defence ratings, and even team motivation/psychology in the form of expert knowledge could lead to improved results. 

\cite{berrar2019incorporating} developed two methods in order to create two feature sets for result prediction - recency features and rating features. Recency feature extraction involved calculating the averages of statistics across the past nine matches, based on four feature groups: attacking strength, defensive strength, home advantage and opposition strength. Rating features were based on performance ratings of each team, updated after each match according to the expected and observed match outcomes, as well as the pre-match ratings of each team. The XGBoost and kNN algorithms were applied to each of these two feature sets. Both of these models were found to perform better on the rating feature set, and the best performance overall was obtained with XGBoost on the rating features - however this result was achieved after the competition. The best model submitted for the competition itself was a kNN applied to the rating features, which achieved accuracy of 51.9\% on the unseen competition test set. It was mentioned that the generally small number of goals per match and narrow margins of victory in soccer meant it is difficult to make predictions based on goals only. The authors concluded that innovative approaches to feature engineering as well as how well domain knowledge can be incorporated into modeling is the key to success for sport result prediction, and are likely more important than the choice of ML algorithm. It is again recognized by the authors that the competition dataset was limited, and that data about various game events (e.g., yellow and red cards, fouls, ball possession, passing and running rates, etc.), players (e.g., income, age, physical condition) and teams or team components (e.g., average height, attack running rate) would help to improve results.

Overall, the competition involving the \textit{Open International Soccer Database} produced some innovative methods and approaches. Notably, researchers in this competition tended to combine some form of ratings-based method with machine learning techniques. Despite having a very large number of matches on-hand in the dataset, all of the studies seemed to obtain accuracies that levelled off at a certain point. This perhaps reflects that having a broad range of predictive features is very important to success in predicting results in sport.

\section{Results \& Discussion}
\label{results-discussion}

In this section we focus on addressing the key research questions that were outlined in the introduction, and in doing so, also provide some critical analysis and discussion of the obtained results.
 
\subsection{Does the evidence from the literature suggest that ANNs produce better performance in terms of predictive accuracy than other ML models?}
\label{sec:candidate-model-discussion}

\begin{figure}[htb]
 \centering\includegraphics[width=1\linewidth]{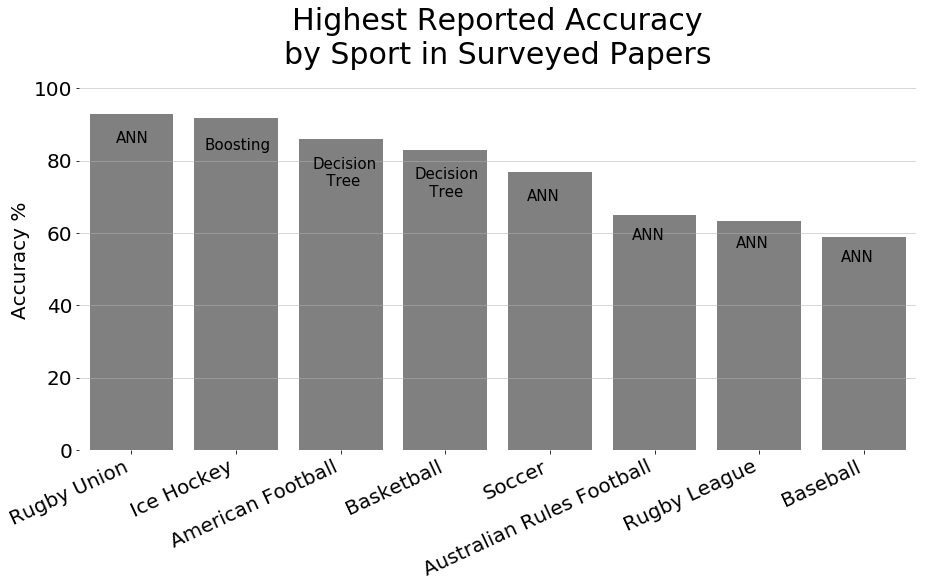}
 \caption{Highest recorded accuracies in research by different sport codes covering the review period. The highest accuracy score is reported with the associated algorithm.}
 \label{highest_reported_accuracy_by_sport}
 \end{figure}

Our research found that the majority of studies (65\%) considered ANNs in their experiments, as seen in Figure \ref{algorithm_usage}. In earlier studies especially, we found that 23\% of the papers considered solely ANNs as their predictive model. The higher propensity for researchers to use ANNs in sports domains has also resulted in the majority of sports papers attributing their highest accuracy to ANNs (Figure \ref{highest_reported_accuracy_by_sport}). It should also be noted that predictive research on two sports in this graph (Rugby League and Australian Rules Football) only considered ANNs as well. Three sports (American Football, Ice Hockey and Basketball), which found their best accuracy performances with alternative algorithms, did in fact also experiment with ANNs; however, alternative algorithms outperformed them.

Therefore, the evidence does not suggest that ANNs have consistently performed better than other ML algorithms in sporting domains. Indeed, the broader ML literature, as well as industry-based applied contexts and ML competitions such as those on Kaggle, do not support blanket statements which would give primacy to ANNs over all other algorithms. 

It is unclear why historically researchers in the sporting domain have tended to display an overwhelming preference for ANNs. ANNs are not straightforward to parameterise optimally and often over-fit, especially in the absence of large enough datasets which tend to be the norm rather than the exception in sporting domains. A further disadvantage of ANNs is that they are not easy to interpret, and are therefore less useful to analysts and team coaches looking to draw out insights, than some other ML models. Most recently, advanced ANNs such as the LSTM, used in the field of Deep Learning, have started to enter the sporting domain \citep{danisik2018football}. Despite low accuracy results from initial research using these algorithms, as datasets increase in size, the accuracy and utilization of these algorithms in this field can be expected to grow.

Decision Tree algorithms were the second most popular techniques (Figure \ref{algorithm_usage}) used across all the papers covered in this review. The appeal of these algorithms is obvious in that they are fast to train and usually do not possess a cumbersome number of tunable parameters. Importantly, they do not generate black-box models, but instead embody varying degrees of interpretability depending on their implementation. This property can offer utility to practitioners beyond just the ability to make predictions, but also in providing insight to coaches, management and athletes. Their widespread use has resulted in both American Football and Basketball reporting their highest accuracy results using CART and Logistic Model Trees (LMT) respectively, as the underlying tree induction algorithms. Other variants of Decision Tree implementations that have been popular in literature include C4.5 \citep{quinlan1993c4} and its corresponding J48 implementation in WEKA \citep{witten1999weka}.

Ensemble methods including various Boosting algorithms, as well as Random Forests, together form a top-three category of techniques used in research in this domain (Figure \ref{algorithm_usage}). Given differing degrees of resilience of this family of algorithms to over-fitting, it is unsurprising that they have been liberally used and have registered top accuracy results in Ice Hockey in a recent study that highlighted the potential of ensemble-based solutions \citep{gu2019game}.

    \begin{figure}[htb]
 \centering\includegraphics[width=1\linewidth]{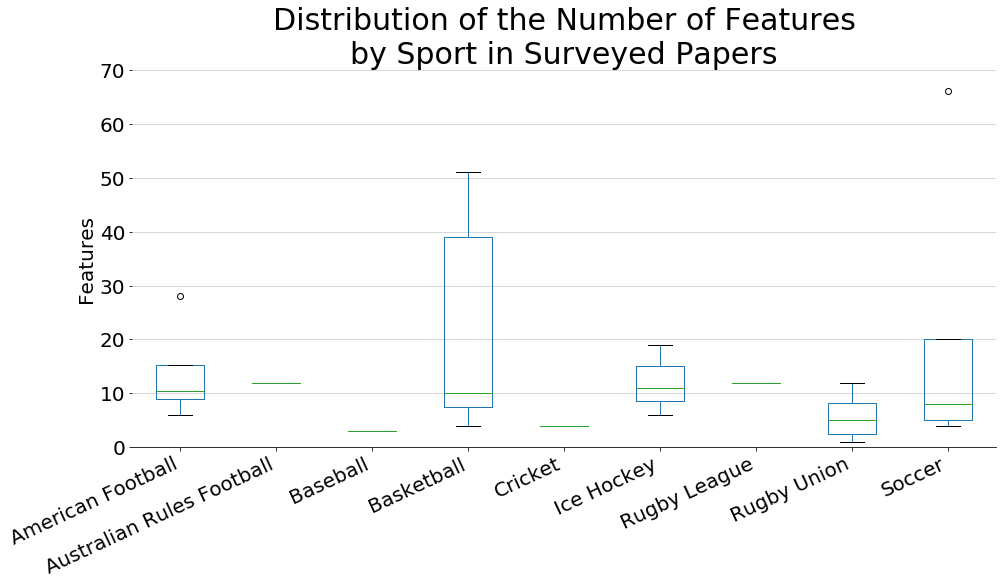}
 \caption{Distribution of the number of features used for machine learning per sport across all surveyed papers. }
 \label{distribution_of_number_of_features}
 \end{figure}
 
Bayesian-based algorithms such as Naive Bayes and Bayesian Networks are one of the most popular sets of techniques used in reviewed research. Though these algorithms are not listed as best performing for individual sports (Figure \ref{algorithm_usage}), they have been found in some comparative studies to offer better accuracies over alternative algorithms \citep{miljkovic2010use,odachowski2012using,joseph2006predicting}. The popularity of Naive Bayes in particular can be attributed to its common usage as a benchmarking algorithm when assessing the learnability of a new problem, which is tied to its ability to generate classifiers that do not over-fit. Bayesian networks were also shown to be useful in incorporating expert knowledge and being effective with minimal data \citep{joseph2006predicting}.

\subsection{What are some of the defining characteristics that can be drawn from successful studies?}
\label{sec:feature-selection-discussion}

In early studies in particular, predictive features were often selected manually by researchers based on their knowledge of the given sport(s). More recently, data-driven methods using various filter-based techniques have become commonplace in successful studies. These have ranged from correlation-based feature subset selection methods \citep{hall1998correlation}, to algorithms such as ReliefF \citep{kira1992practical}, which have a greater contextual awareness and consider the existence of dependencies between features. Others have used feature-importance outputs from machine learning algorithms such as RIPPER \citep{cohen1995fast} in order to inform which features should be retained in the training of final models, while some have used ANN-specific methods including signal-to-noise ratios \citep{bauer2000feature}. Furthermore, sporting experts have also been consulted to select what they consider to be the most important predictive features.

\begin{figure}[tb]
 \centering\includegraphics[width=1\linewidth]{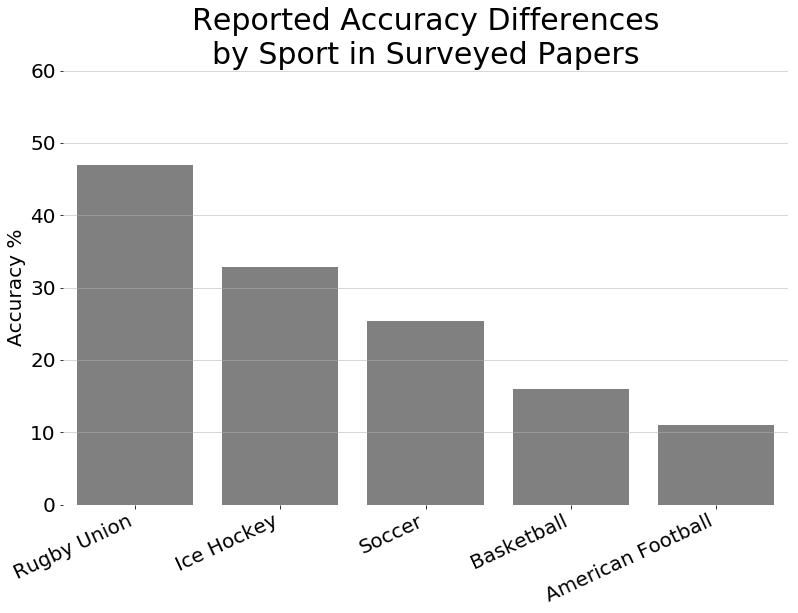}
 \caption{Difference in accuracy percentage points between the lowest and highest predictive results per sport. }
 \label{accuracy_improvements_by_sport}
 \end{figure}

Whilst generally still modest in size, the general expansion of the pool of features used across the various sports over time can be seen in the box-plot of Figure \ref{distribution_of_number_of_features}. Associated with the growth in the size of feature sets, as well as their descriptive capacity, the degree to which accuracy has improved for each sport over time can be observed in Figure \ref{accuracy_improvements_by_sport}. The sports that have seen some of the largest improvements in accuracy over time (Ice Hockey, Soccer and Basketball) are also the sports which have on average exhibited the largest expansions in size of the feature subsets.

In addition to the application of feature selection techniques, studies with a robust experimental process have generally compared a number of different feature subsets. Such subsets could be betting odds, in-play or match-related features, external features, player-level features, expert-selected features, features extracted from pre-game reports, features constructed from ratings, among others. The performance of the various subsets of features can then be compared with each other, as well as with the full set of features.

     \begin{figure}[tb]
 \centering\includegraphics[width=1\linewidth]{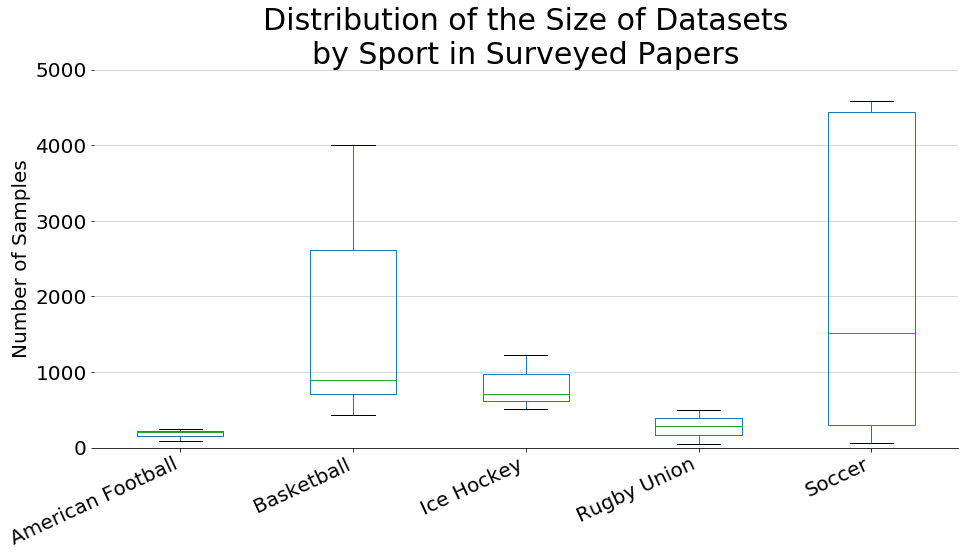}
 \caption{Distribution of the number of samples in datasets used for machine learning per sport across all surveyed papers. }
 \label{distribution_of_number_of_samples}
 \end{figure}
 
The distribution of the size of the datasets and their evolution over time can be seen in Figure \ref{distribution_of_number_of_samples}. It can be observed that generally the datasets have tended to be of a modest size. Sports that have seen an inflation in dataset sizes have been Basketball and, more acutely, Soccer.

\begin{figure}[htb]
 \centering\includegraphics[width=1\linewidth]{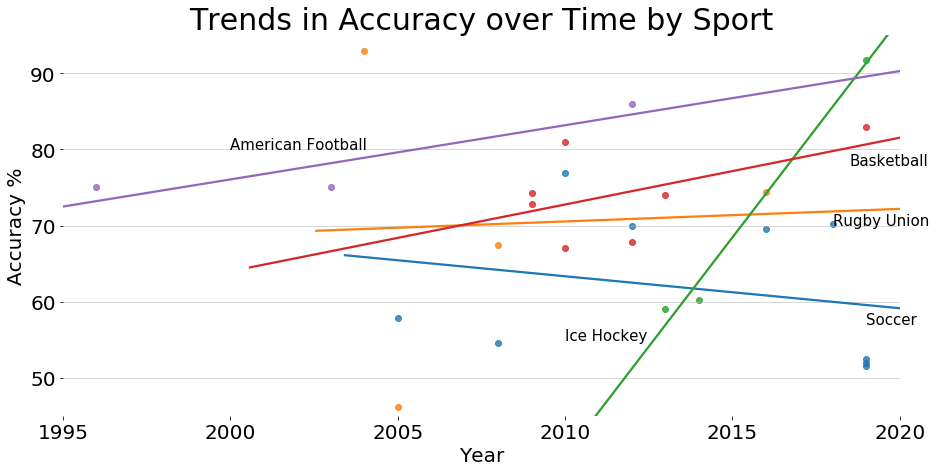}
 \caption{Trends over time in accuracy by each sport. }
 \label{trends_in_accuracy_by_sport}
 \end{figure}

It appears however that having a large dataset in terms of number of matches does not necessarily lead to high accuracy. This was particularly evident in the 2017 Soccer Prediction Challenge Competition, where model accuracies were modest despite having access to over 216,000 matches in their dataset. This particular pattern is depicted in Figure \ref{trends_in_accuracy_by_sport}, where accuracy trends over time are rendered for all sports. Arguably, the growth in both the quantity and quality of features used in each sport (Figure \ref{distribution_of_number_of_features}) has grown more than the size of the datasets, and is thus suggestive of the fact that feature engineering combined with robust feature selection are key drivers of predictive accuracy improvements.

Successful studies have formulated experimental designs that tested a number of different training and validation data splits. For instance, using a certain number of seasons for training a model, while validating the model on another (future) season. If only one season of data was available, some studies trained models on a certain number of competition rounds and then validated on a future round within the same season. Researchers have often varied the split of training seasons/rounds to validation seasons/rounds, and compared the accuracy of models using these different splits. Cross validation (CV) was found to have been used in a number of prior studies. In the sporting domain this can be problematic if the technique is not appropriately modified. CV randomly shuffles instances, meaning that future games might be used to predict past games, and this pitfall was encountered in a number of studies whose predictive accuracy might have been compromised as a consequence.

Comparison of the performance of models predicting the points margin (home team points minus away team points), against a classification approach determining discrete match outcomes (win/loss/draw) was investigated \citep{delen2012comparative, valero2016predicting, danisik2018football}. Both approaches form a distinctive approach to formulating the ML task. On one hand, \cite{delen2012comparative} and \cite{valero2016predicting} found that the classification approach performed better, while \cite{danisik2018football} found that a numeric prediction approach was superior on their dataset. Given these mixed results, we would recommend that, where possible, future researchers perform such a comparison as part of their experimental design.

Researchers have often proposed to apply their predictive models to other sports. However we would emphasize that given each sport has specific features that are unique to it and are associated with outcomes, it is generally not possible to directly apply a trained model to a dataset for a different sport. Rather, it is necessary to go through an new experimental process on a new dataset for the other sport. It is for this reason that we recommend that, in order to approach prediction problems in a structured way, researchers follow an experimental framework such as the CRISP-DM \citep{wirth2000crisp} framework, the Knowledge Discovery in Databases (KDD) framework \citep{fayyad1996advances} or the Sport Result Prediction CRISP-DM (SRP-CRISM-DM) framework \citep{bunker2017machine}. The SRP-CRISM-DM framework is an extension of CRISP-DM that was developed in consideration of the characteristics that are specific to the application of ML to sport result prediction.

  \subsection{Are the results of some sports inherently more difficult to predict than in others?}\label{sec:accuracies_across_sports}

Determining whether predicting outcomes for certain sports is inherently more challenging than in others is difficult to answer conclusively given the existing data. One reason is that sports have received highly imbalanced amounts of attention in the ML literature. This makes it impossible to simply attribute reasons for lower accuracies in sports like Cricket, Rugby League, Baseball and Australian Rules Football to something that is intrinsically more non-deterministic in them, compared to others, when these sports have received minimal research focus (Figure \ref{sport_code_counts}) thus far. However, a little more can arguably be inferred from sports such as Soccer, Basketball, American Football, Rugby Union and Ice Hockey, which have garnered more attention.

Soccer has received the largest share of research, yet its highest recorded predictive accuracy is 78\% (Figure \ref{highest_reported_accuracy_by_sport}) and it comes fifth with respect to accuracy comparisons across all the sports. Rugby Union on the other hand has received limited research focus, yet it ranks the highest of all sports reviewed here, with an accuracy of 93\% \citep{o2004evaluation} being recorded as its best outcome. Interpreting this requires caution. Low scoring sports tend to embody a higher degree of random chance as a determinant of outcomes and this in part explains some of the variation. However, this differential in accuracies can to a large degree be attributed to the characteristics of the sports and the competitive contexts to which the predictive experiments were applied. For instance, Rugby Union is a much smaller global sport than Soccer, being played by fewer nations, while historically being dominated by a handful of them. Understandably then, the best recorded results for Rugby Union originated from a study \cite{o2004evaluation} on the 2003 Rugby World Cup, which exemplified a context with a high degree of predictability\footnote{It should be noted that Rugby Union is becoming less predictable as the gap between low ranked and high ranked nations is narrowing over time, and consequently, more upsets are occurring. This was particularly evident in the reduced predictability of the 2015 Rugby World Cup compared to previous tournaments \citep{o2016predictive}.}. The context for Soccer and competitions from which the results were collected, were diametrically opposite. The results were drawn from both national and international events, where the depth of competition was greater, and which ultimately created conditions for which accurate prediction of outcomes was less deterministic.

Evidence that supports the notion that a great deal of the predictability of sports is naturally determined by the inherent depth of the competitions under observation, rather than on the sports themselves, is supported by Figure \ref{accuracy_improvements_by_sport}, which shows a much higher variability in predictive accuracy for Rugby Union than for Soccer. We can see that in Rugby Union studies that considered national premiership \citep{reed2005development} and international franchise competitions \citep{mccabe2008artificial}, which exhibit a greater degree of equality between teams, the accuracy was correspondingly much lower at 46\% and 74\% respectively.

Additionally, Soccer is a low-scoring sport compared to Rugby Union and this is also a contributing factor to generally lower accuracies in its research. In lower scoring sports, there is a larger element of randomness in outcomes which decreases the performance of predictive models. Associated with this is the higher likelihood of games ending in draws. Given that draws are not improbable, many Soccer researchers have formulated the ML task as a three-class problem (win/loss/draw), rather than a binary-class problem. As the number of classes increase, the learning problem also becomes more difficult and thus accuracies tend to reduce.   

Ice Hockey on the other hand is a counter example even though a binary-class approach was used throughout. The sport is relatively low-scoring although higher scoring than Soccer, yet its most recent and best result (92\% accuracy) has demonstrated considerably higher accuracies in comparison, as well as in comparison to American Football (86\% accuracy) and Basketball (83\% accuracy), which are both high-scoring sports. The research from Ice Hockey, American Football and Basketball, has overwhelmingly been derived from American national league competitions like the NHL, NFL and NBA, which generally exhibit high degrees of competitiveness between the various teams. To some degree this had the effect of controlling for the lopsidedness in expectations of outcomes that can exist in some sports and global competitions. When examining the trends over time in these three sports, some potential patterns emerged to help explain Ice Hockey's unexpected higher accuracies compared to the other two sports. We see that the best performing result in Basketball doubled the number of features that were used in its research over the previous work (74\% accuracy, \cite{loeffelholz2009predicting}) on NBA games. However, the size of the dataset decreased and comparable ML algorithms were used, with standard Decision Tree based algorithms performing the best out of the suite of methods explored. The best performing result in American Football saw an 11 percentage point improvement over the next best result in the NFL \citep{kahn2003neural}, which can be attributed mostly to a threefold increase in the total number of features used in its research (28) from the prior work, since both the total size of the dataset remained essentially the same and the ML algorithms and training procedures did not represent a considerable increase in innovation. On the other hand, the leap from 60\% accuracy \citep{weissbock2014combining} to 92\% accuracy for Ice Hockey can be attributed to multiple factors. The authors used three times as many features, doubled the size of the previous dataset and used more sophisticated algorithms and training approaches in the form of ensembling. 

The points-scoring systems and the manner in which points are attributed to scoring events differ depending on the sport, and this can affect predictability. For instance, goals in Soccer increment the score by one, and there are no other possibilities for a team to increment the score by more than this from any one scoring play. Ice Hockey is similar in this respect. On the other hand, in Basketball, it is possible to increment the score by different amounts from different scoring actions - in particular, via three-point throws, in-circle shots worth two points, and free-throws worth one point. Such differences in score increments according to the scoring action can also be observed in Rugby Union, Rugby League, American Football and Australian Rules Football. Sports that have different possibilities for increments in score have more possible permutations in the final match scores by each team, and therefore final result.

\begin{figure}
\centering\includegraphics{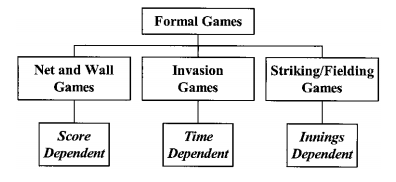}
\caption{Classification of formal games according to \cite{read1992teaching}, retrieved from figure 3 in \cite{hughes2002use}.}
\label{fig:formal_games_types}
\end{figure}

It should be noted that the preceding discussion largely relates to so-called invasion sports (figure \ref{fig:formal_games_types}), which are characterised by a dependence on time, and usually consist of time periods such as halves or quarters. Striking/fielding innings-dependent sports such as Baseball and Cricket have been more lacking in quality research related to sport result prediction, despite being very popular sports. Given its popularity and the fact that the highest predictive accuracy achieved for predicting outcomes is 60\%, Cricket presents a rich ground for further research and accuracy improvements. Baseball could also benefit from additional research given both the small amount of papers that have covered it, as well as the relatively low accuracy achieved (as can be observed in Figure \ref{highest_reported_accuracy_by_sport}).

In summary, though some invasion sports do embody characteristics such as being low-scoring, which makes outcomes harder to predict accurately especially if a multi-class formulation is used, the competitive depth of the types of competitions in which the games take place is also an important factor. In general, matches within sports, competitions or tournaments that are highly competitive, are low-scoring, and have less possible increments in score, and will generally be more difficult to predict. However, the evidence does suggest that these difficulties can to be overcome to a degree when large as well as rich feature sets are used, in conjunction with datasets that have a substantive number of samples, together with the use of cutting-edge training procedures that combine multiple algorithms into robust ensemble-based solutions.   


\subsection{What are the best ways to evaluate accuracy results in this application domain?} \label{sec:benchmark-data-discussion}

Authors in a previous review paper \citep{haghighat2013review} highlighted the inherent difficulties in comparing the results of studies in this application domain. This difficulty arises due to studies usually differing in at least one of the following dimensions: the sport considered, the input data set, the model predictors, the class label, and/or the matches, seasons or competitions that were considered. Part of the challenge lies in the scarcity of available benchmark datasets. The Open International Soccer Database \citep{dubitzky2019open} was compiled recently; however, as was mentioned previously, this particular dataset is limited because it does not include sport-specific features derived from events within matches. Sport related datasets are also becoming increasingly available on websites such as Kaggle.com, which could produce benchmark datasets in future.

Alternatively, there are a number of other ways for researchers to evaluate their experimental results, namely, by comparing results to some baseline measure. Common approaches that have been used in the literature include comparing the predicted outcomes with: \begin{itemize}
      \item Outcomes derived from the match betting odds - the outcome with the lowest betting odds acts as the class label.\footnote{Betting odds have also been shown to be useful for match result prediction even when included as the sole model predictor \citep{tax2015predicting}. However, \cite{hubavcek2019exploiting} pointed out that if the purpose of the model is to generate profit through betting strategies, the betting odds should not be included as a model predictor if one wants to "beat the house".}
    \item Outcomes made by experts on the same matches, e.g., those published online or in newspapers.
    \item A rule that always selects the majority class. In most cases this will predict a home team victory due to the existence of the well-known home-advantage phenomenon.
    \item A randomly selected outcome.
\end{itemize}

Given the specific nature of most datasets in sporting domains, comparisons such as these are useful for researchers to make when reporting their experimental results.

\subsection{What are the common themes of future research directions that can be identified from surveyed papers?} \label{sec:sport-perf-discussion}

\begin{figure}[htb]
 \centering\includegraphics[width=1\linewidth]{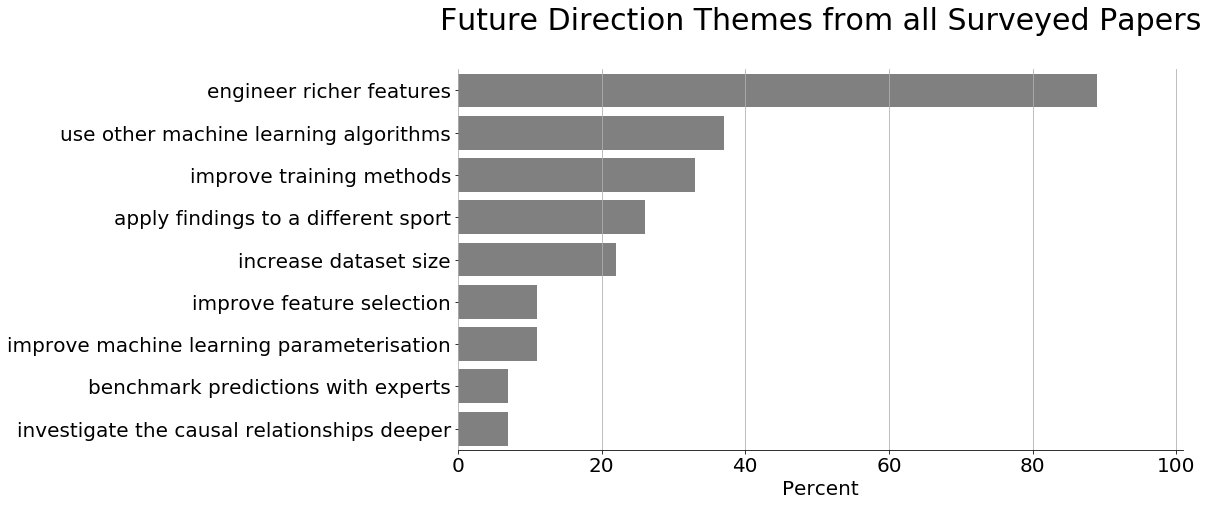}
 \caption{Percent of future direction research themes cited by all papers cited in this study.}
 \label{future_research_direction}
 \end{figure}

Figure \ref{future_research_direction} depicts the percentage of research themes cited as a future direction by papers covered in this study. Only research themes that were cited more than once are shown on the graph. Nearly 90\% of papers listed engineering additional richer features as one of their primary endeavours for future work. This should come as no great surprise, since generating more descriptive and therefore discriminatory features with the help of domain knowledge, is generally accepted as the best strategy for improving predictive accuracy using ML \citep{domingos2012few}. The ability to generate more effective decision boundaries between instances of different classes is to a larger degree determined by richer features then by algorithms of increasing sophistication. To that end, domain expertise will play an important role in crafting more descriptive features. Such domain expertise could be obtained in consultation with coaches or athletes, or potentially from academic literature related to the field of sport performance analysis. Much of the research in this field considers so-called Performance Indicators (PIs), which can be defined as a selection, or combination of action variables that aims to define some or all aspects of a performance, which can be used to assess the performance of an individual, team, or parts of a team, over time \citep{hughes2002use}. PIs essentially form the core of variables that make up sport result prediction models, but are also augmented with features external to sport matches that may have an influence on outcome, e.g., weather, venue, travel, player availability. Sport performance analysts meanwhile are not usually concerned with such external variables because they are usually outside the control of coaches and players. We suggest that there is an opportunity for knowledge transfer from this field, which is likely to be found in the identification and development of new features that act as relevant model predictors. Given that feature engineering has been identified in this review as an area of highest priority for future directions, an increased collaboration between these disciplines may result in meaningful advances. However, the two disciplines do not appear to have reached the point where a significant level of interchange is currently taking place, and researchers are encouraged to expand their collaborations in this respect.

Experimenting with alternative machine learning algorithms was cited as a future undertaking by nearly 40\% of the surveyed studies. Each ML algorithm embodies within it assumptions about characteristics of the problem dataset which may or may not hold, and the extent of the disconnect between the two will also affect the generalizability of different algorithms to varying degrees. A reasonable course of action when an improvement in accuracy is sought, is to investigate different families of algorithms, which is the intention that Figure \ref{future_research_direction} appears to indicate that researchers are heading towards. What is more, non-parametric algorithms have a tendency to overfit, and this is exacerbated on smaller datasets. A quarter of the studies signaled their pursuit to increase the size of their datasets, which in this instance would be the correct course of action for studies that have experienced this difficulty.

Meanwhile, improving training methods by reformulating a ML problem can also have a significant effect on accuracy, which nearly 40\% of the studies intend to explore. The types of training modifications that were cited for future work have included using different combinations of previous season results (where appropriate) and giving greater weight to datasets that are more recent. Some have proposed applying clustering to the datasets before ML, and using alternative algorithms on different datasets.

Given that each sport has very unique characteristics and potential inflection points that can act as markers for winning outcomes, it is surprising that a quarter of the studies plan to apply their ML methodology, including features, to other sporting codes. This seems rather counter-intuitive, given that 90\% of these papers intend to generate more custom-designed features that are more tailored for their respective sporting codes.

Furthermore, it was also unanticipated to see only 10\% of the studies cite improvements in feature selection as an area of pursuit for future research. The problem of over-fitting is amplified when the pool of features is too large with respect to the size of the datasets. Given that most of the available datasets for ML in this field do not tend to be vast, coupled with a strong consensus across all studies to pursue engineering and generation of more features, the problem of over-fitting is likely to plague many studies unless efforts in employing feature selection or dimensionality reduction are afforded equal attention.

\subsection{Limitations}

In this review we have focused on studies that have made use of at least one ML technique. There may be other studies from the statistics, operations research or mathematics literature that may also be relevant or useful for researchers in this field in improving their experimental approach and results which were overlooked here. We encourage researchers to also consult the literature in such fields, which may contain useful insights, particularly if they are studies which relate to the same sport.

The scope of this review also considered only papers related to team sports. Papers related to individual sports (some of which were mentioned in the introduction) may provide useful insights to researchers, particularly if they are looking at incorporating player-level data into their team sport prediction model.

Comparisons between accuracies in studies within the same sport (e.g., in our tabular summaries in section \ref{lit-review}) were made despite these studies having generally used different time-frames/seasons as well as different predictive features. Given the large variability that exists even between studies covering the same sport, there is a limit to the reliability of directly comparing and drawing conclusions from these differences.

While we have discussed match outcome prediction in depth, we have only briefly touched on how prediction of sporting outcomes can be applied for the purpose of coming up with profitable betting strategies. Of course, it is one thing to predict the outcome, but another to come up with a profitable strategy when betting on matches. This includes decisions in terms of whether the betting odd spread is sufficient to warrant the risk of actually placing a bet. Such considerations should be taken into account if one's purpose is to use a predictive model for betting on sport outcomes or events.

Finally, from the perspective of a coach, it is often important how a team plays, not simply the final match result. For instance, a given team could win but still play poorly against a weak opposition or, a team could play well but still lose against a strong opposition. Future work could consider such alternative labels to predict instead of, or in addition to simply win/loss/draw outcomes, e.g., good win, average win, poor win, good loss, average loss, poor loss and so on (as determined by a coach).

\section{Conclusions}
\label{limitations-conclusions}

This paper has provided an in-depth and up to date review of studies that have applied machine learning techniques for the purpose of predicting results in team sports. We analyzed various characteristics of past studies including: the types of algorithms used together with the best performing techniques, the number of features included, and the total number of instances (matches) that authors had available in their dataset.

Our findings suggest that just like in any other field, a wide set of candidate algorithms and ensembles should be used in experimentation in sport result prediction, and that the movement away from just employing one algorithm such as ANN, should continue to take place.
Engineering rich sets of features is important, more so than having a large number of instances.
It is worthwhile experimenting with different subsets of features and making robust comparisons between them and their performance against full feature supersets. 
Exploring the comparison of numeric prediction models (predicting points margin), classification models (win/loss/draw), and even labels based on a coaches' evaluation of performance (e.g. excellent/average/poor) warrants further investigation. Given the characteristics of specific sports and their datasets, it is recommended that experimental process frameworks are followed (e.g., KDD, CRISP-DM, SRP-CRISP-DM). Since k-fold cross-validation shuffles instances randomly, it should be altered in a manner that generates order-preserved folds, so as to prevent future matches being used to predict past matches, e.g., using a season-by-season or round-by-round training-validation split. Benchmark datasets for particular sports are still limited in their availability. However, model predictions can be compared to predictions from betting odds, expert predictions, and baselines such as a rule that always select a home-side victory, or a randomly selected match outcome.

Sports and competitions have varying degrees of inherent competitiveness and thus predictability which are determined by whether they are high or low scoring, whether there are different score increments for each scoring play as well as different probabilities of draws occurring.

Lastly, there is an opportunity for knowledge transfer from sport science and sport performance analysis to improve sports result prediction, particularly in the identification of match-related features, while Baseball and Cricket are popular team sports that could benefit from more quality research.

%
%
%

\bibliographystyle{model5-names}
\bibliography{biblio}

\end{document}